\documentclass[letterpaper, 10 pt, conference]{ieeeconf}  

\usepackage[top=54pt,bottom=54pt,left=54pt,right=54pt]{geometry}

\usepackage{amssymb, amsmath, amsthm}
\usepackage{mathtools}
\usepackage{physics}
\usepackage{siunitx}

\usepackage{cite}
\usepackage{graphicx}
\usepackage{multirow}
\usepackage{makecell}

\usepackage{blindtext}
\usepackage{algorithm}
\usepackage{algpseudocode}
\usepackage{hyperref}
\usepackage{cleveref}   

\usepackage{verbatim}

\usepackage{makecell}
\usepackage{tikz}
\usetikzlibrary{fit}

\IEEEoverridecommandlockouts                              

\title{\LARGE \bf
Estimating Force Interactions of\\Deformable Linear Objects from their Shapes
}

\author{Qi Jing Chen$^{1}$, Shilin Shan$^{1}$, Timothy Bretl$^{2}$, and Quang-Cuong Pham$^{3}$
\thanks{*Github repo: \url{https://github.com/qj25/ds2f} (Videos: \url{https://youtu.be/_jDbKWxA19w})}
\thanks{$^{1}$Nanyang Technological University, School of Mechanical and Aerospace Engineering, $^{2}$University of Illinois Urbana-Champaign,
$^{3}$Eureka Robotics, Singapore
	}%
}

\newcommand{\cmmnt}[1]{\ignorespaces}
\begin{document}

\maketitle

\begin{abstract}
	
This work introduces an analytical approach for detecting and estimating external forces acting on deformable linear objects (DLOs) using only their observed shapes. In many robot-wire interaction tasks, contact occurs not at the end-effector but at other points along the robot’s body. Such scenarios arise when robots manipulate wires indirectly (e.g., by nudging) or when wires act as passive obstacles in the environment. Accurately identifying these interactions is crucial for safe and efficient trajectory planning, helping to prevent wire damage, avoid restricted robot motions, and mitigate potential hazards. Existing approaches often rely on expensive external force-torque sensor or that contacts occur at the end-effector for accurate force estimation. Using wire shape information acquired from a depth camera and under the assumption that the wire is in or near its static equilibrium, our method estimates both the location and magnitude of external forces without additional prior knowledge. This is achieved by exploiting derived consistency conditions and solving a system of linear equations based on force-torque balance along the wire. The approach was validated through simulation, where it achieved high accuracy, and through real-world experiments, where accurate estimation was demonstrated in selected interaction scenarios.

\end{abstract}


\section{Introduction}
Force feedback of robotic interaction with wires can be useful for trajectory planning or implementing safety conditions during collisions~\cite{schlechter2001manipulating,lee2015learning,nakagawa2018real,suberkrub2022feel,zhong2023regressor}. In many such scenarios, the robot does not directly grasp or manipulate the wire -- for example, when a robotic arm nudges a wire into place. In other cases, wires serve merely as passive elements or environmental obstacles that the robot must navigate around. For the case of a robot arm, a common robotic manipulator for wires, force sensors are generally mounted on the end-effector. This allows for interaction forces of an end-effector gripper with a wire to be easily recorded. The problem arises when robot-wire interactions do not happen at the end-effector but instead at other locations along the robot arm. These interactions must be considered because they could lead to restricted or potentially dangerous movement of the robot. Too much tension in the wires could lead to breakages of the wires in the environment. Understanding these interaction forces will allow robots to plan their trajectory based on expectations of how a wire would react to its immediate movement.

For this reason, we introduce an algorithm which estimates the forces on a wire based on its shape. From here we will refer to the wire as an elastic rod. Our work derives consistency conditions, based on force torque balance equations, which determine if sections of an elastic rod in static equilibrium belong to the same undisturbed section where no external interaction forces are present between them. With the section classification and knowledge on the internal stiffness torques of the elastic rod based on the discrete elastic rods (DER) model~\cite{bergou2008discrete}, directions and magnitudes of external forces on the rod can be solved for. Due to the indeterminate nature of the problem, the algorithm proceeds by either assuming zero external torque and solving for the force positions, or by assuming that forces are applied at the midpoint of each section and solving for the resulting external torque. A solution to obtain both torques applied and force positions is by detecting collision points along the wire in the real image and identifying their positions visually, although this method will not be discussed in our work. We observe that our approach is closely related to existing proprioceptive sensor-based force and collision estimators~\cite{wahrburg2017motor,wang2023active,shan2023fine}. However, prior work primarily applies these methods to robotic arms, where accurately localizing contact forces and estimating their magnitudes remains challenging. In contrast, our method transfers this idea to the wire, leveraging the internal torques of an elastic rod as a proxy for proprioceptive sensing within a similar estimation framework. The inherent discretization of the wire allows for more accurate force localization and estimation.

\begin{figure}[t]
	\centering
	\setlength{\fboxsep}{0pt}
	\begin{tikzpicture}
		\node[inner sep=0pt] at (2.0, -1.31) {\includegraphics[trim=0 1.2cm 0 1.2cm,clip,width=0.2088\textwidth]{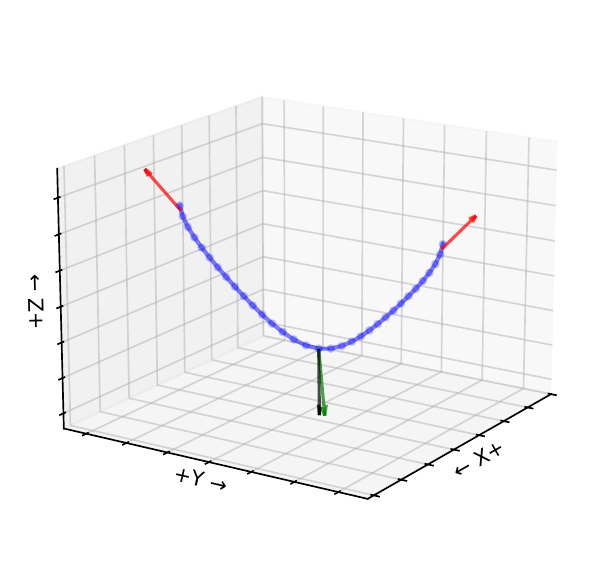}};
		\node[inner sep=0pt] at (2.0, 1.3) {\fbox{\includegraphics[trim=2.0cm 3.0cm 2.0cm 2.55cm,clip,width=0.225\textwidth]{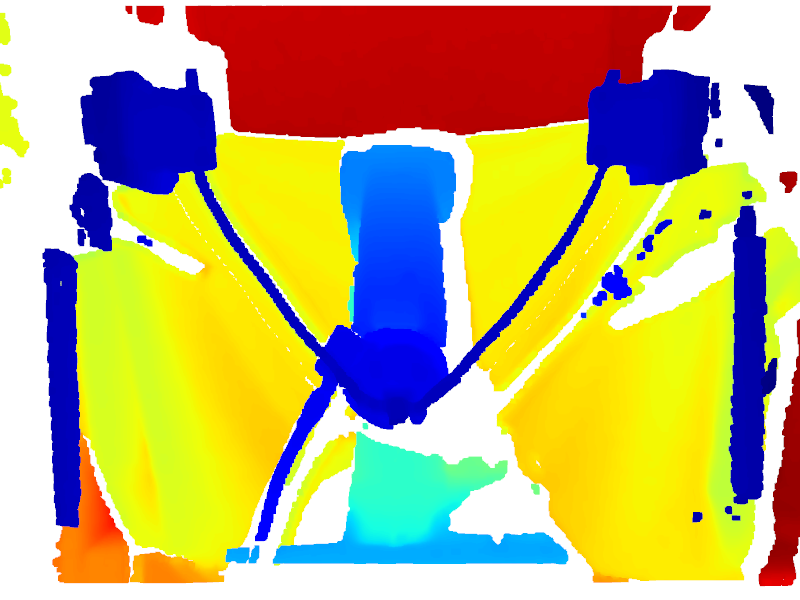}}};	
		\node[inner sep=0pt] at (-2.01, -1.31) {\fbox{\includegraphics[trim=3cm 0 3cm 0,clip,width=0.225\textwidth]{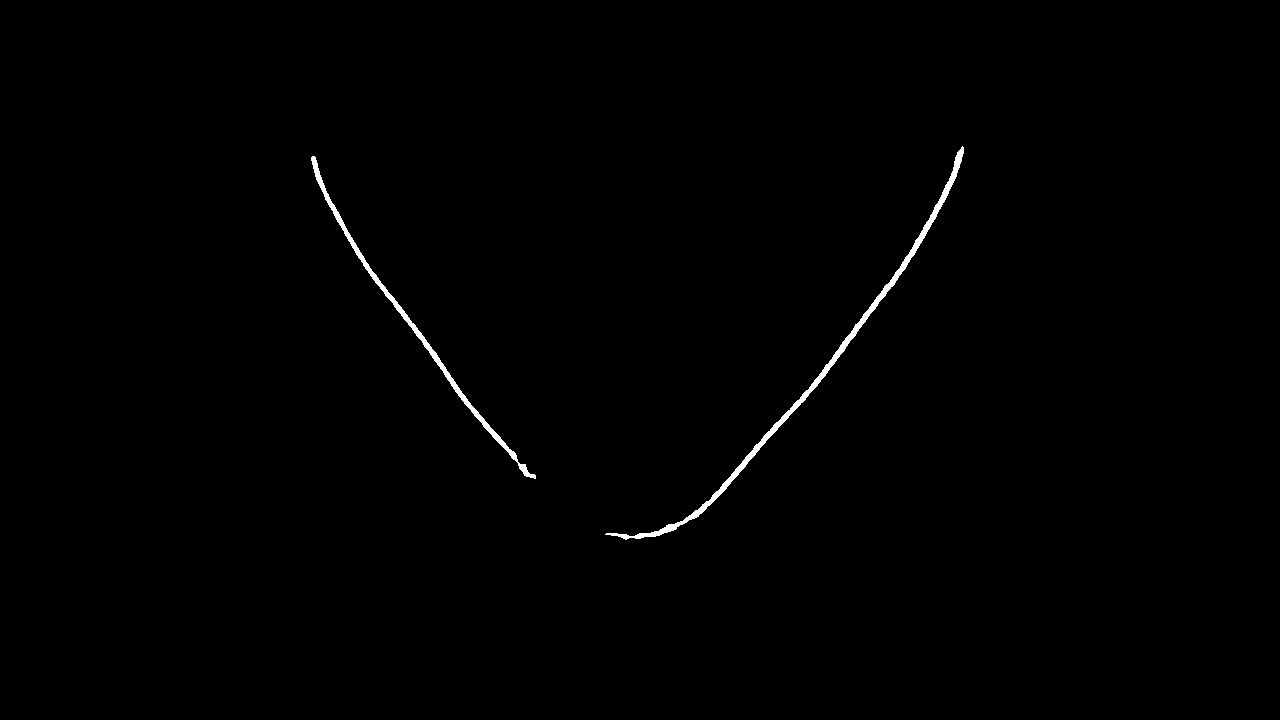}}};
		\node[inner sep=0pt] at (-2.01, 1.3) {\fbox{\includegraphics[trim=3cm 0 3cm 0,clip,width=0.225\textwidth]{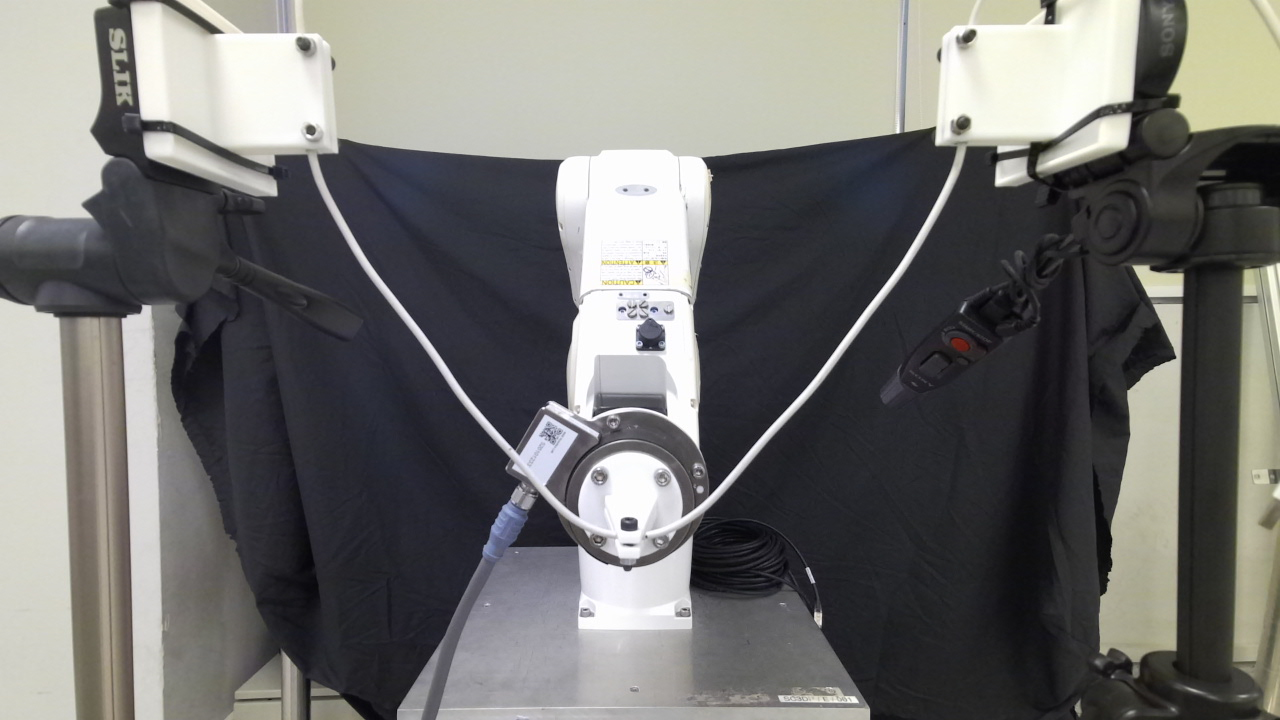}}};
	\end{tikzpicture}
	
	\caption[Overall Force Estimation Process]{Overall Force Estimation Process. Rowwise from top left to bottom right: Real image of experiment, depth information, segmentation mask of wire, and final smoothed wire shape with actual and estimated forces (arrows: red - estimated end-clamp force, black - actual force, green - estimated external force).}
	\label{Fig:s2f_overallpipeline}
\end{figure}

\subsection*{Contribution and organization of the paper}
The contributions of this paper are twofold. First, we formulate a set of novel consistency conditions to estimate the positions of external interactions along an elastic rod in static equilibrium (\Cref{Section:estimatingpos}). Prerequisite assumptions are also stated to ensure the problem is not underdetermined.

Second, using the positions of interactions, parameters of the force and torque can be solved for using a system of linear equations derived from the force-torque balance in each discretized rod piece (\Cref{Section:estimatingdisturb}). Due to the nature of the problem, one can either solve for the external torque applied or the exact positions of force application, the former requiring an assumption of force application being at the center of mass of the discrete piece, and the latter with the assumption that there is no external torque applied at that discrete piece.


\section{Related Works}
Robotic manipulation of Deformable Linear Objects (DLOs) increasingly emphasizes force sensing and control to overcome limitations such as unknown object properties and visual occlusions. Robots operating in complex or unstructured environments require sophisticated capabilities to safely interact with their surroundings, particularly concerning detecting physical contact and understanding interaction forces.

\subsection{Force Control for DLO Manipulation}
Foundational work investigated characteristics in force signals for detecting contact state transitions between a DLO and rigid obstacles~\cite{schlechter2001manipulating}. Research reveals that the set of all static equilibrium configurations for a Kirchhoff elastic rod constitutes a smooth manifold of finite dimension that can be explicitly parameterized by moments and forces at the elastic rod base~\cite{bretl2014quasi}, and its free configuration space is path-connected~\cite{borum2015free}. This significantly simplifies and enables the construction of gripper paths for manipulation planning of DLOs. Complementary methods focus on estimating DLO shape or contact information primarily from force/torque sensing, achieving real-time 3D shape estimation of elastic rods via a discretized Kirchhoff elastic rod model with gravity compensation \cite{takano2017real, nakagawa2018real}, or by keeping the DLO under tension for contact inference and primitive execution \cite{suberkrub2022feel}. Another method is to learn force-based manipulation skills from demonstrations for variable-impedance control, enabling tasks like knot-tightening despite challenges in capturing accurate force profiles \cite{lee2015learning}. Furthermore, regressor-based model adaptation offers an online adaptation law for unknown DLO deformation parameters, facilitating model-based force control without vision feedback for open-loop shape control in quasi-static scenarios \cite{zhong2023regressor}.

\subsection{Contact Detection and Force Estimation}
One direct solution for force sensing involves using external sensors, such as distributed tactile 'robot skin' sensors, to detect contact intensity and location across the robot's body and drive it through obstacles~\cite{novak1991capacitance,albini2021exploiting}. However, integrating such extensive and accurate sensing equipment is often expensive~\cite{nuelle2017force}. To that end, it was found that proprioceptive sensors of the robot can be used for contact detection~\cite{de2005sensorless,haddadin2008collision, cho2012collision}. Two main issues faced when using proprioceptive sensors to detect collisions are the system dynamic modeling inaccuracies and the noise from proprioceptive sensors~\cite{haddadin2017robot}. These problems affect the accuracy of the estimated external joint torques and could lead to false positives in contact detection.

Simpler robot models are generally successful in force estimation~\cite{6224977}, some integrating visual information as a tool~\cite{lee2018interaction}. Difficulties in modeling the non-linear dynamics of the whole robot arm has led to the utilization of a model-based Kalman filters with motor signals~\cite{wahrburg2017motor}, an Extended Kalman Filter torque fusion method~\cite{wang2023active}, and a model-free leaning-based neural network approach~\cite{shan2023fine}. Such methods require accurate calibration in the event of modifications to the robot structure which affect its dynamics~\cite{gaz2017payload,shan2024fast}. Although these works showed some success in estimating the direction, magnitude, and position of external forces, they require additional assumptions such as having only one external contact point~\cite{han2019collision}, or that the contacts are applied sequentially~\cite{manuelli2016localizing}, due to the underdetermined nature of the problem. The handling of numerous contacts is still a problem that demands attention. This idea of contact detectability has been discussed in-depth~\cite{pang2021identifying}, in which small motions that most effectively falsify spurious contact positions has been found. Vision and depth sensing has been utilized in a GPU parallel processing algorithm to effectively determine contact points on a robot arm and react accordingly~\cite{magrini2017human}.


\section{Problem Definition}
\label{Section:ws}
In the following, we use the terms ‘elastic rod’ and ‘wire’ interchangeably. We define a rod piece as a discrete section between consecutive nodes of a discretized rod. A section is then defined as a set of consecutive rod pieces that share one or more properties. We categorize pieces of a discretized elastic rod into two section types (s-types): Undisturbed (UD) and Disturbed (D). A UD section is defined as a set of consecutive discretized rod pieces to which no external disturbances are applied. A D section is defined as a set that has at least one disturbance applied to one or more of its pieces. An elastic rod is discretized into $n = n_{\text{UD}} + n_{\text{D}}$ pieces where $n_{\text{s-type}}$ is the total number of s-type pieces (s-type being either UD or D). $N = N_{\text{UD}} + N_{\text{D}}$ is the total number of sections and $N_{\text{s-type}}$ is the number of s-type sections. Each edge in the discretized rod is defined as $\mathbf{e}^{i}=\mathbf{x}_{i+1}-\mathbf{x}_{i}$ where $\mathbf{x}_{i}$ is the Cartesian position of the $i-\text{th}$ piece. Each D section has a set of external force-torque, $\{\mathbf{f}^j, \boldsymbol{\tau}^j\}$, where $0 \leq j < N_{\text{D}}$. Theoretically, our method should be able to detect all force disturbances on an elastic rod provided it fulfills the following condition:
\begin{itemize}
	\item Elastic rod is in static equilibrium or in sufficiently quasistatic motion. This forms the basis of our method and allows us to derive the equations required.
	\item Rod must be sufficiently discretized such that there are at least 3 discrete undisturbed pieces between consecutive sections of external disturbances. This ensures that a necessary and sufficient condition can be formed to categorize pieces.
	\item The 3 undisturbed pieces identified must not be parallel to ensure the problem is not underdetermined. To understand this intuitively, shape of a completely taut wire does not visibly change in response to variations in external force, making it difficult or even impossible to visually infer the force distribution.
	\item Rod is behavior closely follows the elastic rod theory used. This ensure that the categorization and estimation of forces is accurate.
\end{itemize}

\subsection{Elastic Rod Theory}
Before the estimation of external forces and torques can occur, one would have to provide an accurate elastic rod model to compute the internal torques within the system. For our work, we have chosen to use the discrete elastic rods (DER) theory~\cite{bergou2008discrete}, which has been implemented as a plugin in MuJoCo~\cite{todorov2012mujoco}. Additionally, we estimate the material properties of the wire with a simple parameter identification pipeline~\cite{chen2025accuratesimulationparameteridentification}.

\section{Formulation}
\label{Section:forms2f}
The total torque on piece $i$ of the discretized rod is
\begin{equation}
	\boldsymbol{\tau}_{total} - c\boldsymbol{\omega} = \mathbf{I} \boldsymbol{\alpha}
\end{equation}
where $\mathbf{I}$ is the second area moment of inertia, $\boldsymbol{\alpha}$ is the angular acceleration vector, $c\boldsymbol{\omega}$ is the damping torque and $\boldsymbol{\tau}_{total}$ is the overall torque on the system. When the rod is in static equilibrium, $\boldsymbol{\alpha} = 0$, $c\boldsymbol{\omega} = 0$.
\begin{equation}
	\boldsymbol{\tau}_{total} = \left(\mathbf{e}^{i} \times \mathbf{F}^i\right) + \left(\mathbf{a}^i \times \mathbf{f_{c}}^i\right) + \boldsymbol{\tau_c}^i + \mathbf{c}^i = 0,
\end{equation}
where $\mathbf{F}^i = \sum_{j\in G_i} \mathbf{f}^j$ such that $G_i$ is the set of all external forces belonging to D sections after piece $i$ (i.e., D sections containing pieces with index $> i$). This means that $\mathbf{F}^k = \mathbf{F}^l$ if $k$ and $l$ belong to the same UD section. $\mathbf{f_{c}}^i$ and $\boldsymbol{\tau_{c}}^i$ are respectively the external force and torque on piece $i$ and are equal to $0$ if $i$ belongs to an UD section. $\mathbf{a}^i = \mathbf{p}_i - \mathbf{x}_i = r\mathbf{e}^i$ where $\mathbf{p}_i$ is the point of application of $\mathbf{f_{c}}^i$, and $r \in [0, 1] \subset \mathbb{R}$. $\mathbf{c}^i$ is the stiffness torque applied on piece $i$ by its adjacent pieces (not to be confused with unbold $c$). Note that the effects of gravity is easily included by adding the torque effects of gravity from the side which $\mathbf{F}_R$ is acting (i.e., adding $\left(\left(n-i-0.5\right)\mathbf{e}^i \right) \times \mathbf{wpp}$ to $\mathbf{c}^i$ where $\mathbf{wpp}$ is a 3-vector describing the weight per piece of the discretized wire).

\subsection{Identifying Positions of Disturbances}
\label{Section:estimatingpos}
Suppose two pieces $i$ and $i+1$ belong to the same UD section, the following equations must be consistent.
\begin{align}\label{eq:consist2}
	&\mathbf{e}^{i} \times \mathbf{F}^i = -\mathbf{c}^i \\
	&\mathbf{e}^{i+1} \times \mathbf{F}^{i+1} = -\mathbf{c}^{i+1}
\end{align}
Knowing $\mathbf{F}^i = \mathbf{F}^{i+1}$, we arrive at the first consistency condition (condition A) which states that $\mathbf{e}^i\cdot\mathbf{c}^{i+1} + \mathbf{e}^{i+1}\cdot\mathbf{c}^{i} = 0$. This is a necessary condition to conclude that these adjacent pieces belong in the same UD section. To check for sufficiency, we investigate the case of piece $i$ belonging to an adjacent D section. 
\begin{align}\label{eq:necctest}
	&\mathbf{e}^{i} \times \mathbf{F}^i + \mathbf{a}^i \times \mathbf{f_{c}}^i + \boldsymbol{\tau_c}^i = -\mathbf{c}^i \\
	&\mathbf{e}^{i+1} \times \mathbf{F}^{i+1} = -\mathbf{c}^{i+1}
\end{align}
Using triple vector product rule, we get 
\begin{equation}
	\mathbf{f_{c}}^i \cdot \left(\mathbf{e}^{i+1} \times \mathbf{a}^i \right) + \mathbf{e}^{i+1}\cdot\boldsymbol{\tau_c}^i = -\left(\mathbf{e}^i\cdot\mathbf{c}^{i+1} + \mathbf{e}^{i+1}\cdot\mathbf{c}^{i}\right).
\end{equation}
Knowing that $\text{RHS} = 0$ and $\mathbf{e}^{i+1} \times \mathbf{a}^i \neq 0$ (linear independence condition), we find that the consistency condition can be fulfilled in two cases. The first is when $\mathbf{f_{c}}^i = 0$, which confirms that pieces $i$ and $i+1$ belong to the same UD section. The second is when $\mathbf{f_{c}}^i \cdot \left(\mathbf{e}^{i+1} \times \mathbf{a}^i \right) = 0$. Since $\mathbf{a}^i \parallel \mathbf{e}^i$, this means that condition A can be fulfilled even when the two pieces belong to different sections as long as $\mathbf{f_{c}}^i$ lies in the plane spanned by $\mathbf{e}^{i}$ and $\mathbf{e}^{i+1}$, and $\boldsymbol{\tau_c}^i \perp \mathbf{e}^{i+1}$. Therefore, condition A is not a sufficient condition.

To establish a sufficient condition for pieces to belong to the same UD section, we must check the consistency of 3 adjacent pieces.
\begin{align}
	\text{Given	} &\mathbf{A}_i = \left[\left[\mathbf{e}^i\right]_\times \quad \left[\mathbf{e}^{i+1}\right]_\times \quad \left[\mathbf{e}^{i+2}\right]_\times\right]^T \quad 
	\\ \text{and} \quad &\mathbf{C}_i = -\left[\mathbf{c}^{i} \quad \mathbf{c}^{i+1} \quad \mathbf{c}^{i+2}\right]^T\text{,}
	\\ \text{ where } &\left[\mathbf{e}\right]_\times \text{ is the skew symmetric of 3-vector } \mathbf{e}. \notag
\end{align}
The consistency of the above equation according to a manually designed threshold is the second consistency condition we term as condition B -- a necessary and sufficient condition to conclude that the pieces belong to the same UD section. We solve for $\mathbf{F}^{*} = \underset{\mathbf{F}}{\operatorname{argmin}} \norm{\mathbf{A}\mathbf{F} - \mathbf{C}}_2^2$ using least squares.
\begin{equation}
	\mathbf{F}^{*} = \mathbf{A}^{\!\!+}\mathbf{C}
\end{equation}
where $\mathbf{A}^{\!\!+} = \mathbf{V}\boldsymbol{\Sigma}^{\!+}\mathbf{U}^T$ is the pseudoinverse of $\mathbf{A}$ arrived at from the singular value decomposition (SVD) of $\mathbf{A} = \mathbf{U}\boldsymbol{\Sigma}\mathbf{V}^T$.

\subsection{Estimating External Disturbances}
\label{Section:estimatingdisturb}
After we have classified the discretized wire pieces into their D and UD sections, we solve for external disturbances.

\subsubsection{Estimating External Forces}
It is interesting to note that it is not possible to solve for forces with only one UD piece $\mathbf{e}^{i} \times \mathbf{F}^i = \left[\mathbf{e}^i\right]_\times \cdot \mathbf{F}^i = -\mathbf{c}^i$, as the skew-symmetric of a 3-vector is maximally rank 2, resulting in an underdetermined problem. As such a minimum of 2 pieces are required. The solving and classification processes are combined, and $\mathbf{F}_R$ is calculated as the average of all the $\mathbf{F}^{*}$ for that UD section. 

As the wire consists of alternating D and UD sections, the force on section $D_j$ can be calculated as
\begin{equation}
	\mathbf{f}^j = \mathbf{F}_R^k - \mathbf{F}_R^l + \mathbf{f}_g
\end{equation}
such that $k$ and $l$ belong to consecutive UD sections separated by $\text{D}_j$, $k<l$, and $\mathbf{f}_g^j = n_{\text{D}_j} \times \mathbf{wpp}$ is the force contribution due to gravity.

\subsubsection{Estimating External Torques or Force Positions}
There are three possible solutions following this. First, we either use a known position of force application for the given D section to compute torque on it.
\begin{equation}
	\boldsymbol{\tau}_{\text{D}_j} = \left(\mathbf{e}^j_\text{D} \times \mathbf{F}^j_\text{D}\right) + \left(\mathbf{a}^j_\text{D} \times \mathbf{f}^j\right) + \mathbf{c}^j_\text{D} + \boldsymbol{\tau}^j,
\end{equation}
such that $\mathbf{e}^j_\text{D} = \mathbf{x}_k - \mathbf{x}_l$ where $k$ and $l$ are the pieces on the two ends of section $\text{D}_j$ and $k>l$. $\mathbf{a}^j_\text{D} = \mathbf{p}_j - \mathbf{x}_l$ where $\mathbf{p}_j$ is the point of application of the external force $\mathbf{f}^j$. $\mathbf{c}^j_\text{D}$ is the stiffness torque applied on $\text{D}_j$ by the UD pieces just adjacent to section $\text{D}_j$ (gravity contributions can be added to this term). Assuming static equilibrium $\boldsymbol{\tau}_{\text{D}_j} = 0$, we can now solve for $\boldsymbol{\tau}^j$.

Second, we can assume that torque acting on the D section is 0 ($\boldsymbol{\tau}^j = 0$) and compute the exact point of external force application on the D section ($\mathbf{p}_j$) under the assumption of static equilibrium.

Third, force is assume to be applied at the mass center of the D section. No torque is computed. The results from our real-world experiments utilize this final solution, as the scope of our work is limited to force prediction. Estimating torque would require capturing internal twist from RGB-D data -- a problem that remains an open challenge in the field.


\section{Simulation Experiments}
\begin{figure}[!htbp]
	\centering
	\setlength{\fboxsep}{0pt}
	\begin{tikzpicture}
		\node[inner sep=0pt] at (0.0, 0.0) {\includegraphics[trim=0 0cm 0 0cm,clip,width=0.43\textwidth]{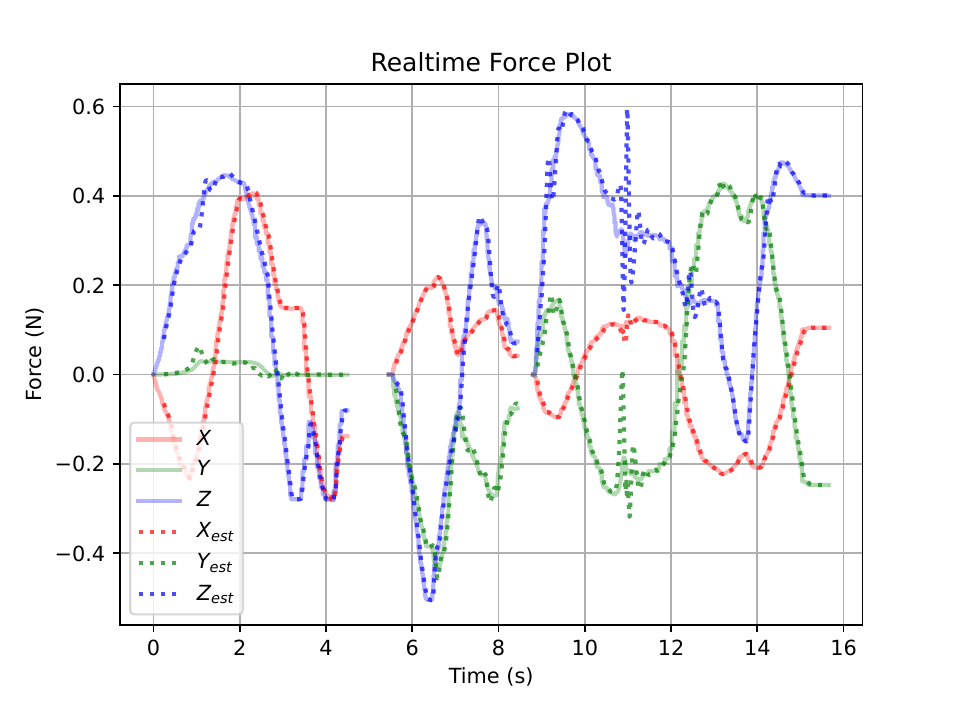}};	
		\node[inner sep=0pt] at (0.0, -5.5) {\includegraphics[trim=0cm 0 0cm 0,clip,width=0.43\textwidth]{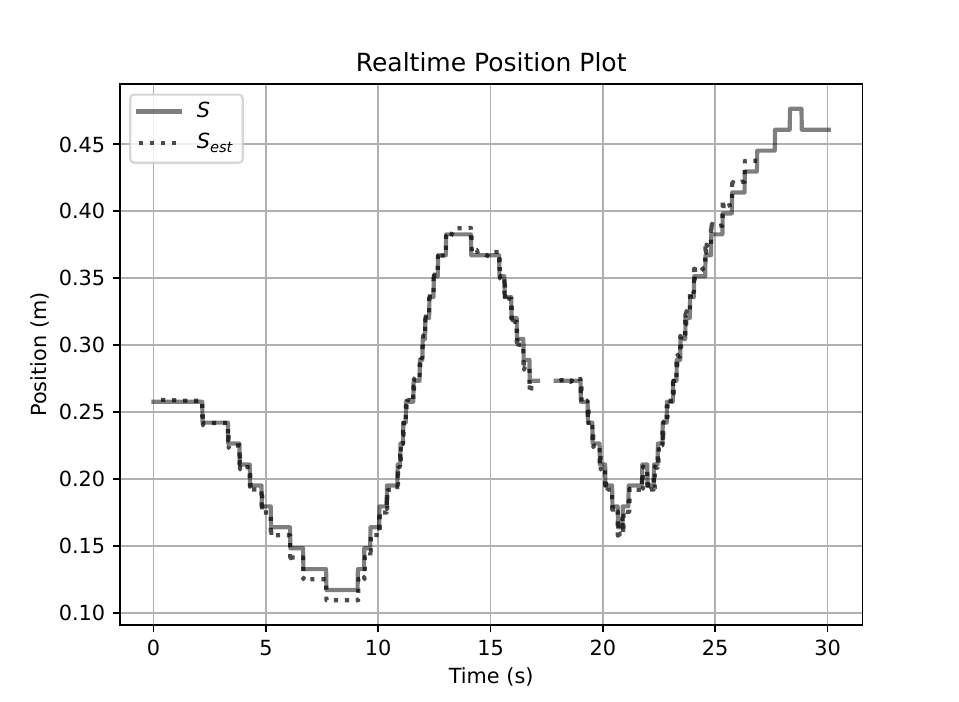}};
		\node[inner sep=0pt] at (0.0, 5.0) {\fbox{\includegraphics[trim=20cm 12cm 23cm 7cm,clip,width=0.35\textwidth]{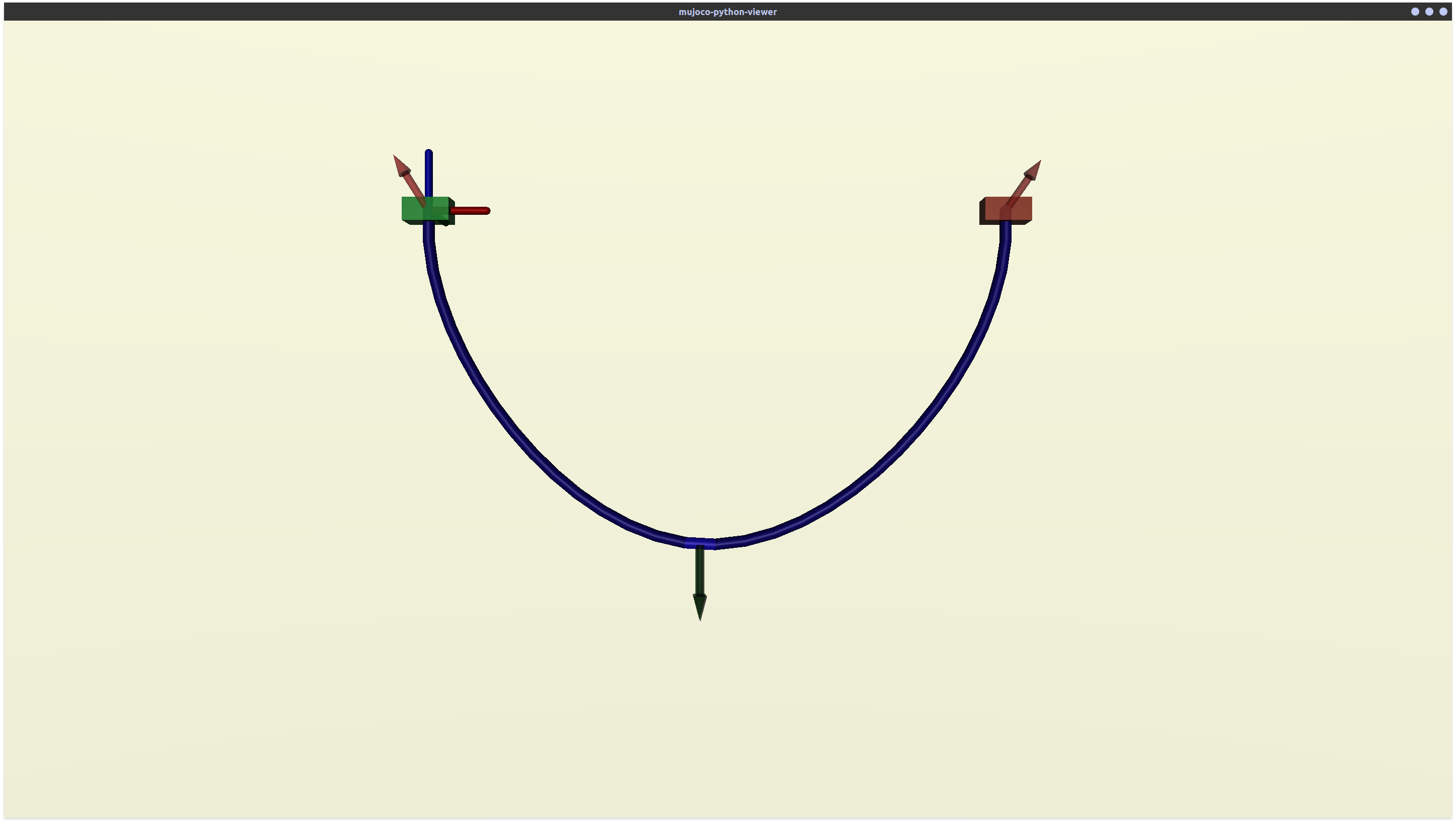}}};
		
		\node[font=\large] at (-1.25, 2.35) {\textbf{P1}:};
		\node[font=\large] at (-1.35, -3.15) {\textbf{P2}:};
	\end{tikzpicture}
	
	\caption[Force Estimation Simulation Experiments]{Simulation experiments for force estimation. The top image shows a screen capture of the wire with a force applied through the interactive user visualization window. The actual (black) and estimated (green) forces are shown (overlapping). Two simulation experiments were carried out: \textbf{P1} shows the force estimation for varying force magnitude and direction on disturbances at the center of the wire length. \textbf{P2} shows the estimation of force position when the point of force application was varied. The solid and dotted lines are the actual and estimated data, respectively. Video of the experiments found here: \url{https://youtu.be/_jDbKWxA19w}.}
	\label{Fig:s2f_simexp}
\end{figure}

A wire was clamped at two points $\SI{50}{cm}$ apart along the wire length. The clamps were held $\SI{30}{cm}$ apart horizontally and the excess wire length was left to dangle. Perfect knowledge on wire stiffness values was assumed. The environment used an adapted DER model~\cite{chen2025accuratesimulationparameteridentification} to simulate bending and twisting behaviors and compute internal stiffness torques. The algorithm used only the wire’s pose from the simulation as input and predicted the resulting external interactions. The wire twist could be captured through a call to the API, a capability which is lacking for real experiments. Because the force estimation from shape was relatively fast, it could be run in real time. Forces were input by the user through an interactive simulation viewer. A video of the simulation experiments can be found at \url{https://youtu.be/_jDbKWxA19w}. Results are plotted in \Cref{Fig:s2f_simexp}. As expected from the assumption of quasistatic equilibrium, force estimation performed better when there was less wire movement. Theoretically, force estimation becomes infeasible when the wire undergoes large motions (effectively becoming an underdetermined problem when too many perceived external forces must be predicted). In spite of this, we visually observed that force estimation was able to perform well even in the presence of significant wire motion.


\section{Physical Experiments}
\subsection{Experimental Set-up}
The set-up followed the simulation experiments. A Denso VS-060 robot arm was fitted with a force-torque sensor at its end-effector along with a custom wire clamp attachment. The robot arm was used to manipulate the wire first at its midpoint and then at a $\SI{7}{cm}$ offset along the wire length from its center. Actual force readings were recorded with the force-torque sensor and filtered using a Chebyshev filter. Due to the limitation of capturing internal twist from visual and depth information, our results assumed a zero-twist configuration of the rod, focusing solely on force prediction and not torque estimation. Although gripper movement would introduce external torque on the wire, this did not affect force prediction, as the force prediction and torque estimation parts of the pipeline were separate. The result was still a prediction of the external force from the gripper on the wire.

The wire shape was obtained through the pipeline presented in~\cite{zhaole2023robust}, leveraging GROUNDED-SAM, a zero-shot image segmentation framework, which we found to be both easy to implement and accurate, with the drawback of longer computational times (overall detection pipeline takes $\SI{8}{s}$ per 720p frame on an NVIDIA GeForce RTX 3060 Ti GPU). The wire was discretized into 60 pieces after smoothing and interpolated into 30 discrete pieces for further processing. For DLO smoothing, we used a trade-off between the error from node displacement due to smoothing ($p_{\text{smooth}}$) and the decrease in elastic potential energy in the rod ($E_{\text{decrease}}$). As long as $J_{i} > J_{i-1}$ where $i$ is the time step and $J = E_{\text{decrease}} - m_p*p_{\text{smooth}}$, smoothing continued. In this work, we applied time stepping through the DER model implemented in MuJoCo and included a plugin state which provides the potential energy in the rod. The same model was used to compute internal torques in the rod.

\subsection{Results}
\begin{figure}[!htbp]
	\centering
	\setlength{\fboxsep}{0pt}
	\begin{tikzpicture}
		\def\ysep{2.8}  
		\def\xsep{2.3}
		\def\textsep{-1.60}
		\def\lsize{0.225}
		\def\rsize{0.225}
		\def\xcoord{-1.0}
		\def\ycoord{0.3}
		
		\node[above=2pt, font=\scriptsize] at (-\xsep, -0*\ysep+\textsep) {(-5, 0, 0)};
		\node[above=2pt, font=\scriptsize] at (-\xsep, -1*\ysep+\textsep) {(0, 0, -1)};
		\node[above=2pt, font=\scriptsize] at (-\xsep, -2*\ysep+\textsep) {(0, 0, 8)};
		\node[above=2pt, font=\scriptsize] at (-\xsep, -3*\ysep+\textsep) {(0, 1, 0)};
		\node[above=2pt, font=\scriptsize] at (-\xsep, -4*\ysep+\textsep) {(-7, 0, 7)};
		\node[above=2pt, font=\scriptsize] at (-\xsep, -5*\ysep+\textsep) {(0, 2, 7)};
		
		\node[inner sep=0pt] at (-\xsep, 0*\ysep) {\fbox{\includegraphics[width=\lsize\textwidth]{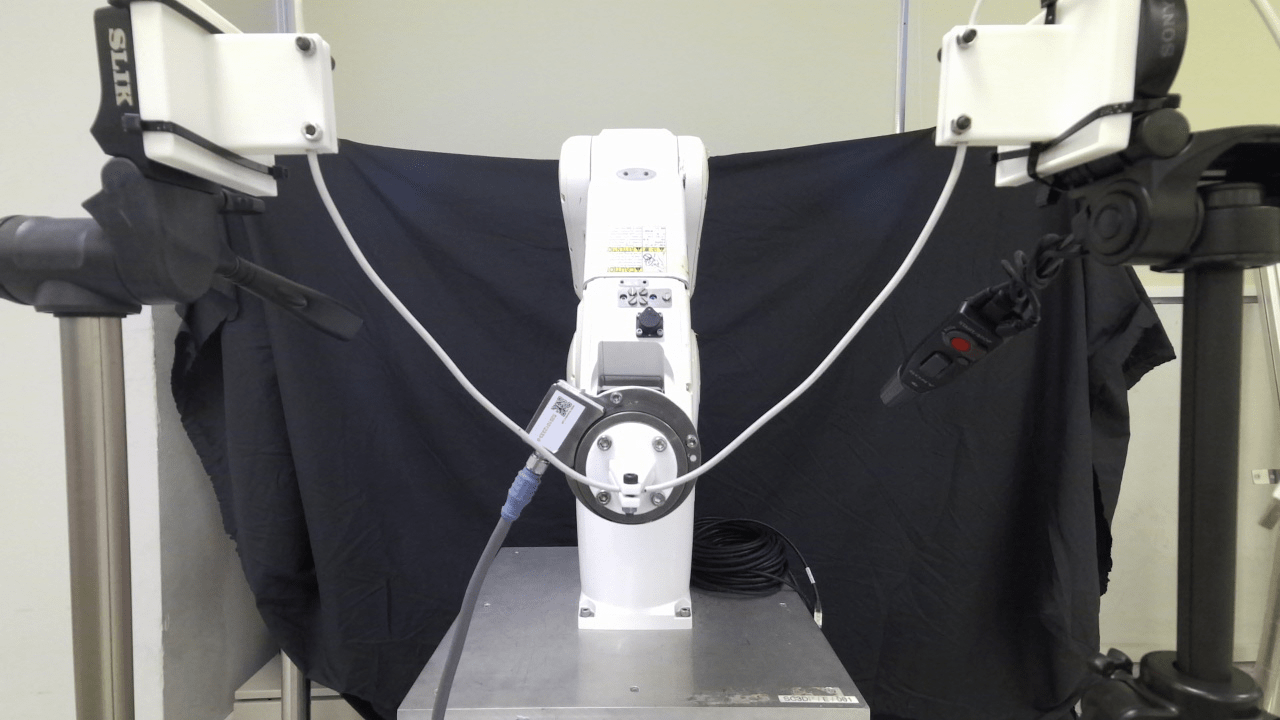}}};
		\node[inner sep=0pt] at (\xsep, 0*\ysep) {\includegraphics[trim=0 1.4cm 0 1.4cm,clip,width=\rsize\textwidth]{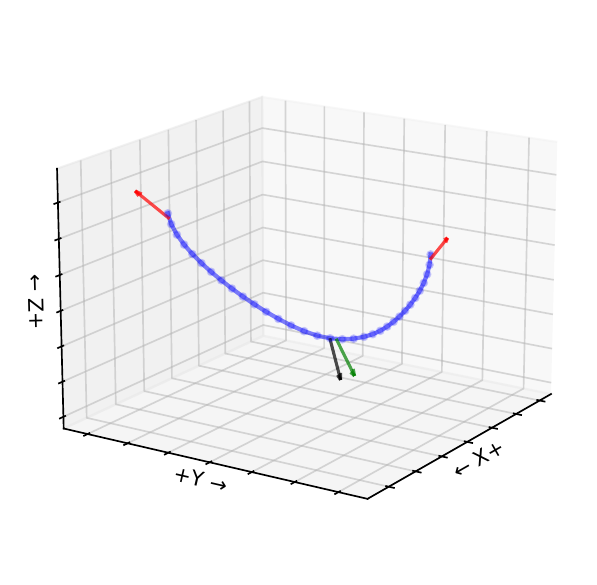}};
		
		\node[inner sep=0pt] at (-\xsep, -1*\ysep) {\fbox{\includegraphics[width=\lsize\textwidth]{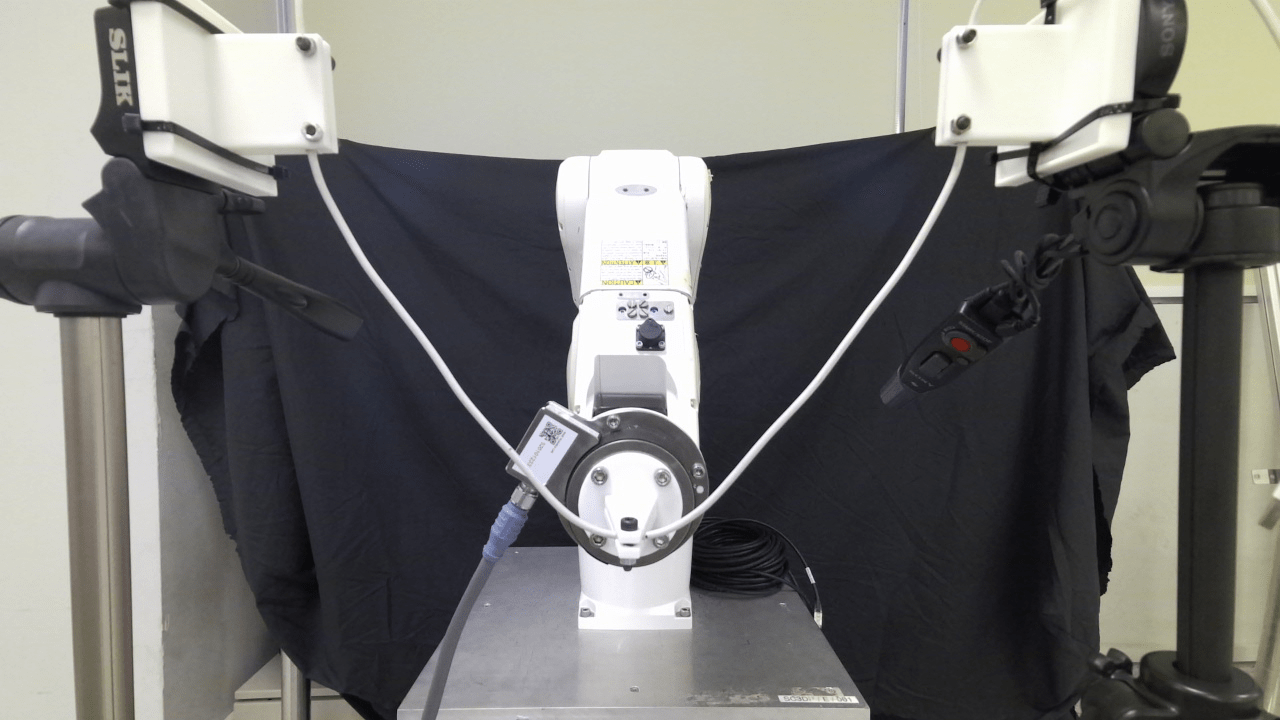}}};
		\node[inner sep=0pt] at (\xsep, -1*\ysep) {\includegraphics[trim=0 1.4cm 0 1.4cm,clip,width=\rsize\textwidth]{pics/exp1/plt_000002.pdf}};
		
		\node[inner sep=0pt] at (-\xsep, -2*\ysep) {\fbox{\includegraphics[width=\lsize\textwidth]{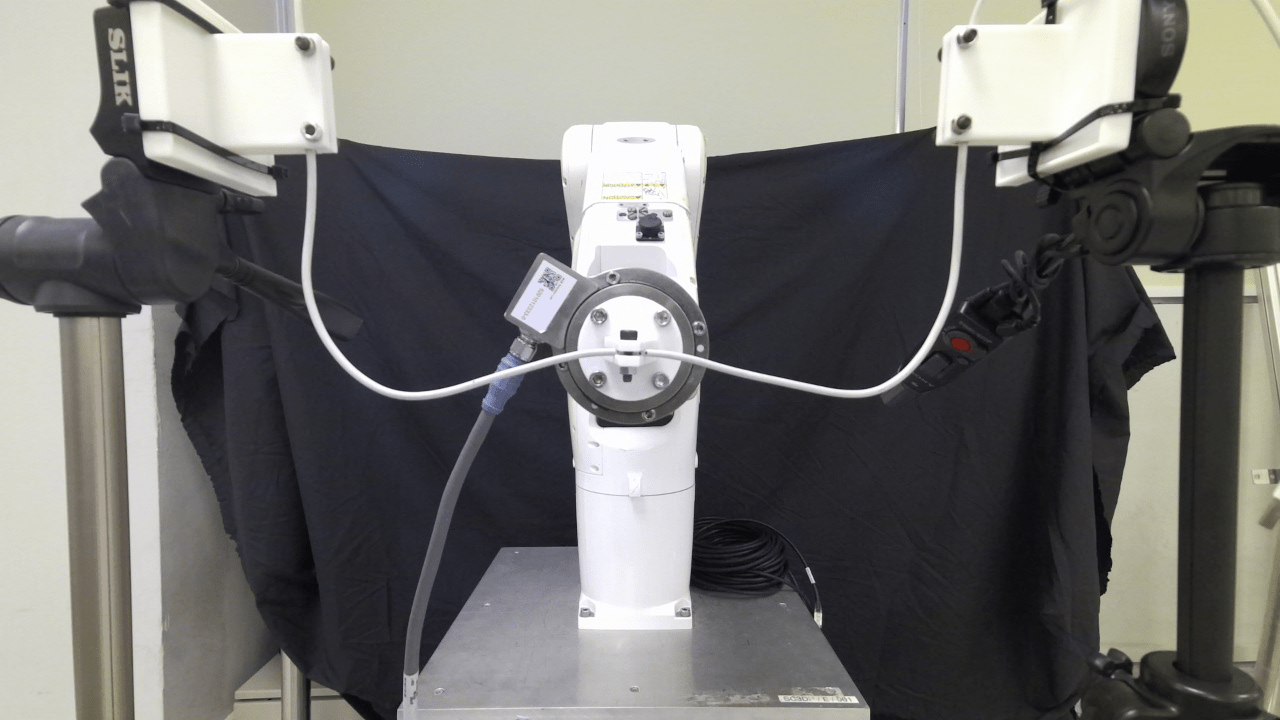}}};
		\node[inner sep=0pt] at (\xsep, -2*\ysep) {\includegraphics[trim=0 1.4cm 0 1.4cm,clip,width=\rsize\textwidth]{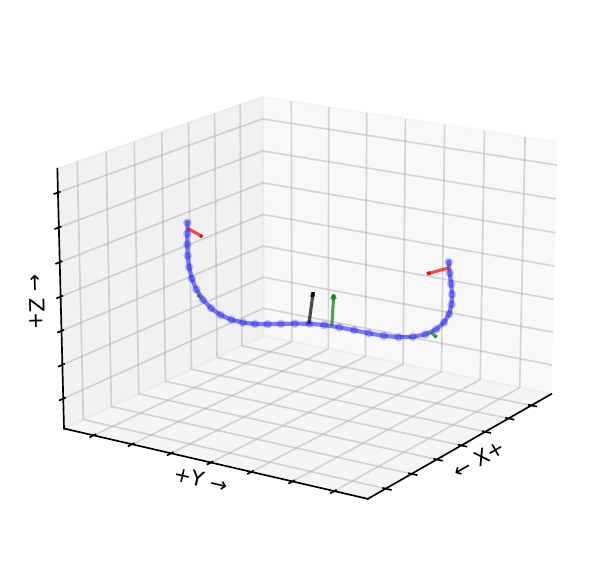}};
		
		\node[inner sep=0pt] at (-\xsep, -3*\ysep) {\fbox{\includegraphics[width=\lsize\textwidth]{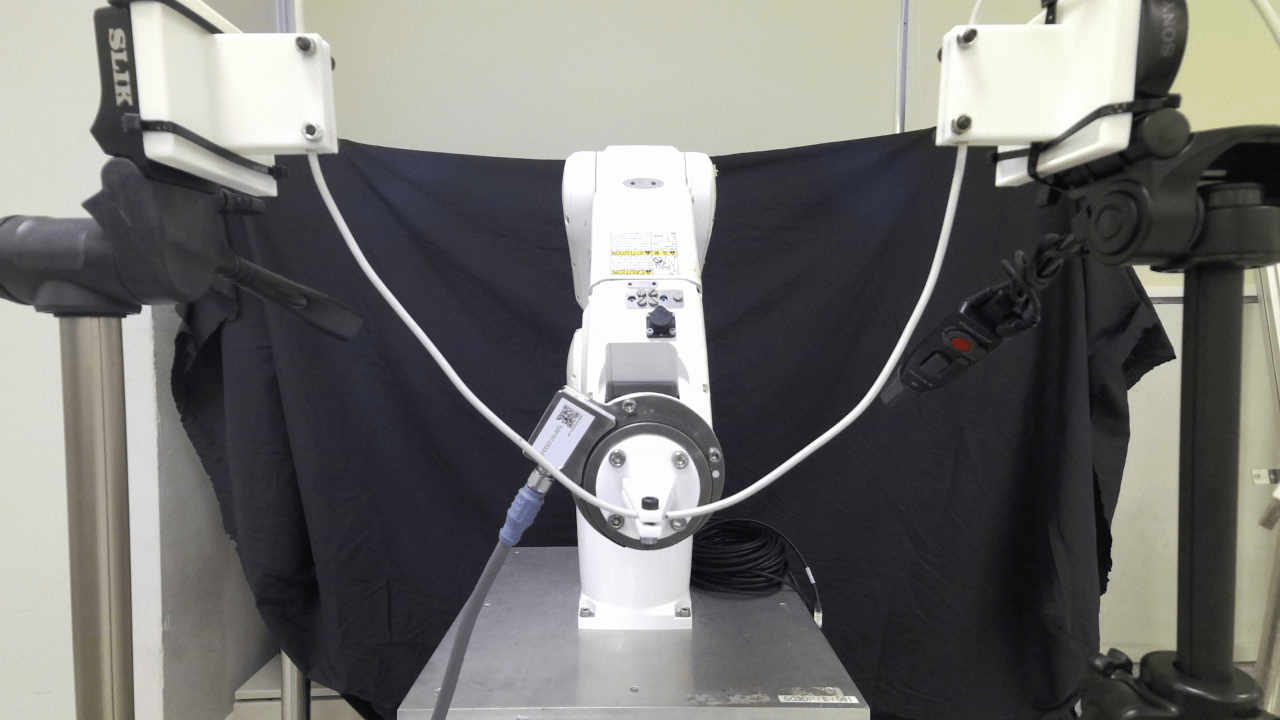}}};
		\node[inner sep=0pt] at (\xsep, -3*\ysep) {\includegraphics[trim=0 1.4cm 0 1.4cm,clip,width=\rsize\textwidth]{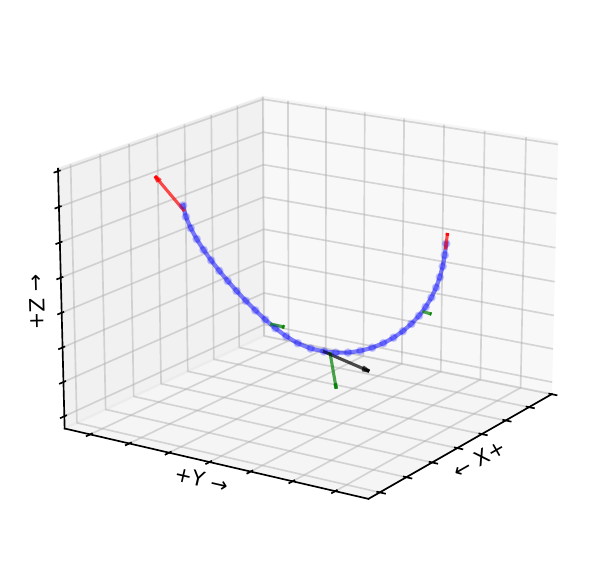}};
		
		\node[inner sep=0pt] at (-\xsep, -4*\ysep) {\fbox{\includegraphics[width=\lsize\textwidth]{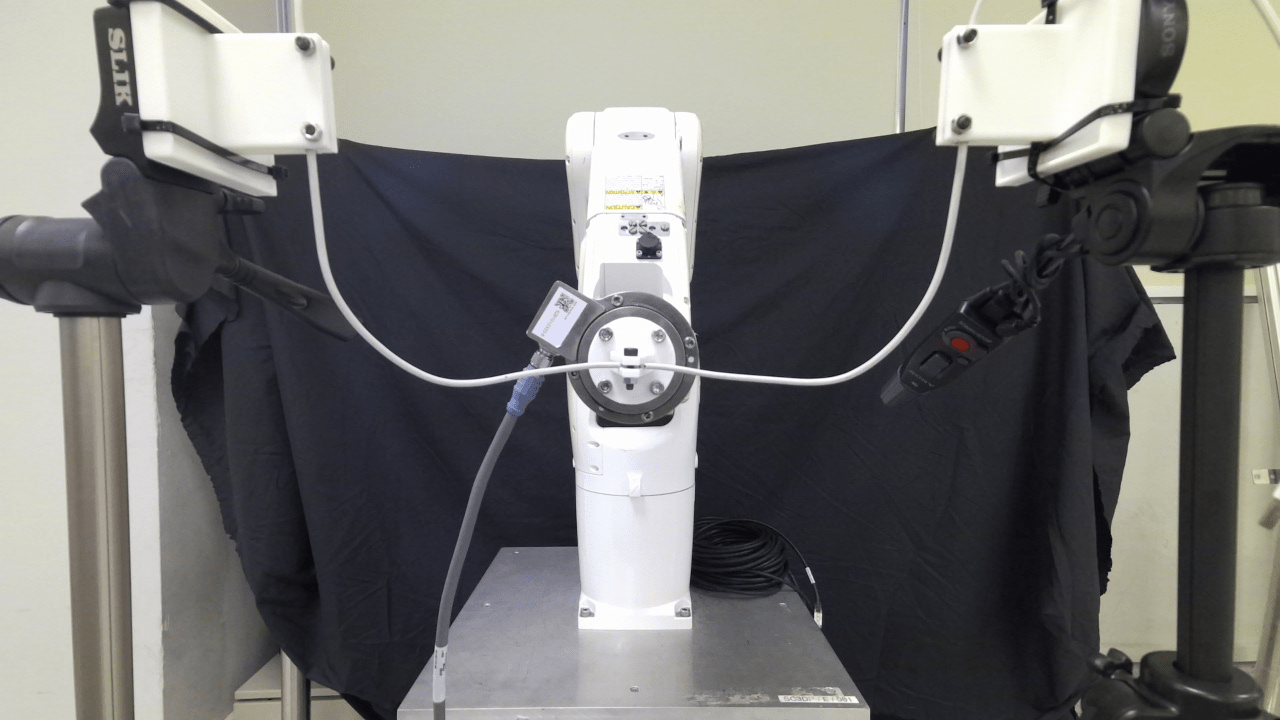}}};
		\node[inner sep=0pt] at (\xsep, -4*\ysep) {\includegraphics[trim=0 1.4cm 0 1.4cm,clip,width=\rsize\textwidth]{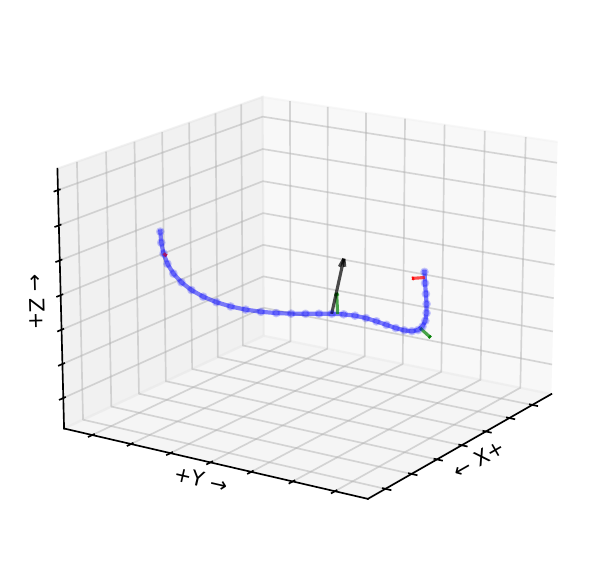}};
		
		\node[inner sep=0pt] at (-\xsep, -5*\ysep) {\fbox{\includegraphics[width=\lsize\textwidth]{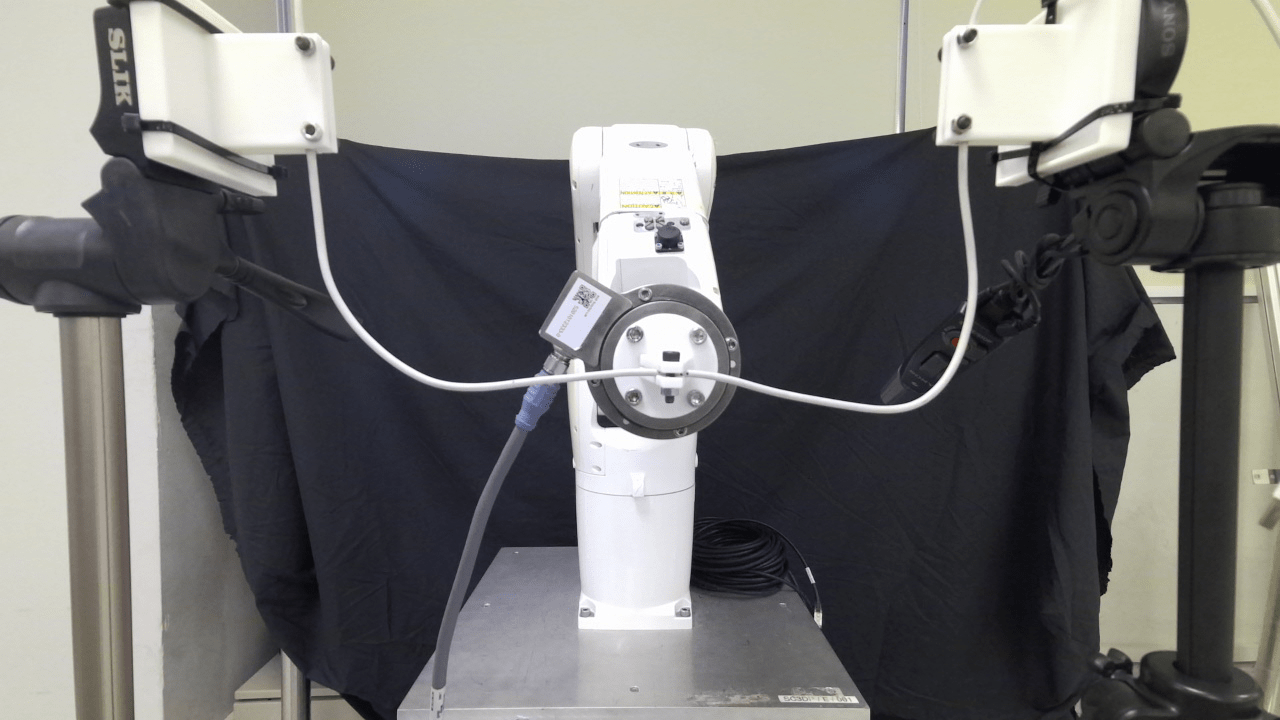}}};
		\node[inner sep=0pt] at (\xsep, -5*\ysep) {\includegraphics[trim=0 1.4cm 0 1.4cm,clip,width=\rsize\textwidth]{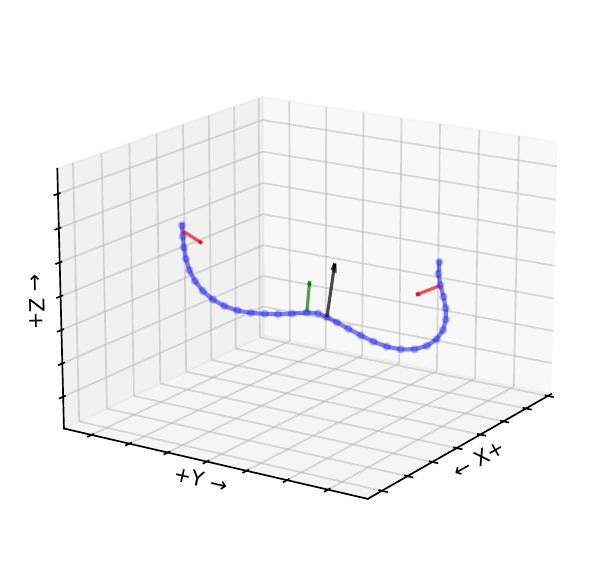}};
		
		\node[
		draw=gray,
		thick,
		fill=white,
		fill opacity=0.75,
		text opacity=1,
		text width=1.3cm,
		minimum height=1.2cm,
		align=center
		] (custom) at (\xcoord+0.2,\ycoord-0.95)
		{};
		
		\draw[->, thick] (\xcoord,-1.2+\ycoord) -- (0.5+\xcoord,-1.2+\ycoord) node[below right] {$\mathbf{y}$}; 
		\draw[->, thick] (\xcoord,-1.2+\ycoord) -- (\xcoord,-0.7+\ycoord) node[above left] {$\mathbf{z}$};  
		

	\end{tikzpicture}
	
	\caption[Visual Results of Force Estimation for Centered Displacement]{Visual results of force estimation for experiment \textbf{A} where the wire was clamped at both ends and attached to the robot end-effector at its center. The robot end-effector was fitted with a force-torque sensor and moved into 6 different positions. The left column shows the real experiment images (coordinates axes shown in first image) with Cartesian displacement of the grasped point (below), and the right column shows the smoothed wire shape along with the actual and estimated force vector (arrows: red - estimated end-clamp force, black - actual force, green - estimated external force). Note that the accuracy of end-clamp forces are not analyzed.}
	\label{Fig:s2fexpA_vis}
\end{figure}

The experiments were split into two main parts. Coordinate frame of the experiment was defined as shown in plots of \Cref{Fig:s2fexpA_vis}. For the first part (\textbf{A}), the wire was clamped to the robot end-effector at its center ($\SI{25}{cm}$ along its length from the left end) in its equilibrium position and moved into 6 different positions with pure translational motion (orientation remains constant). Position displacement from the neutral positions are shown in \Cref{Fig:s2fexpA_vis}.
These positions were selected for variety and the distance was chosen to ensure the condition of non-parallel discrete pieces was fulfilled as much as possible (to avoid an underdetermined problem). The actual and estimated force vectors for experiment \textbf{A} are shown in \Cref{Fig:s2fexpA_vis}. Vectors within each plot are shown to scale relative to each other, but their scales are not consistent across different plots. 

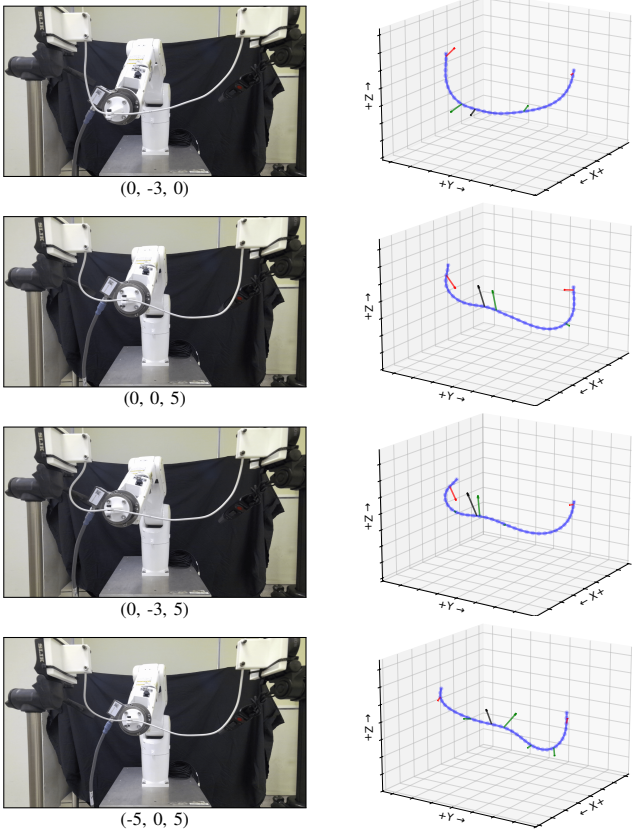
\begin{figure}[!htbp]
	\centering
	\setlength{\fboxsep}{0pt}
	\begin{tikzpicture}
		\def\ysep{2.8}  
		\def\xsep{2.3}
		\def\textsep{-1.6}
		\def\lsize{0.225}
		\def\rsize{0.225}
		
		\node[above=2pt, font=\scriptsize] at (-\xsep, 0*\ysep+\textsep) {(0, -3, 0)};
		\node[above=2pt, font=\scriptsize] at (-\xsep, -1*\ysep+\textsep) {(0, 0, 5)};
		\node[above=2pt, font=\scriptsize] at (-\xsep, -2*\ysep+\textsep) {(0, -3, 5)};
		\node[above=2pt, font=\scriptsize] at (-\xsep, -3*\ysep+\textsep) {(-5, 0, 5)};
		
		\node[inner sep=0pt] at (-\xsep, 0*\ysep) {\fbox{\includegraphics[width=\lsize\textwidth]{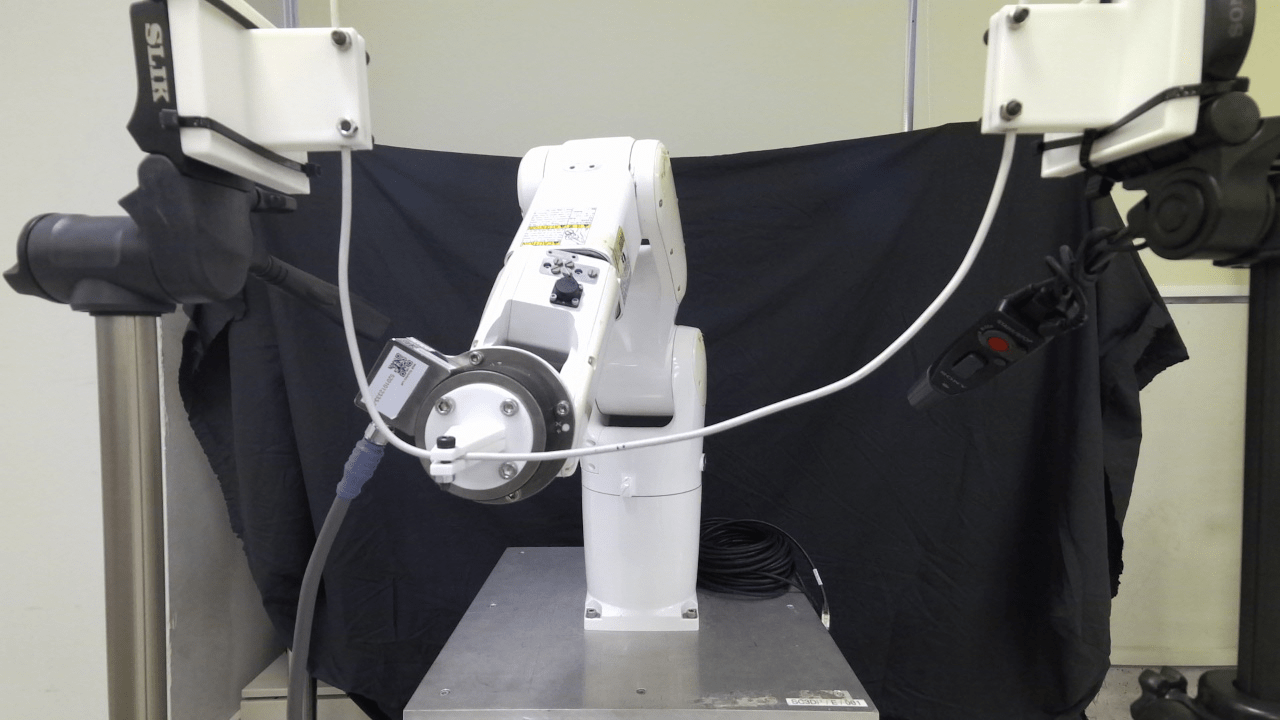}}};
		\node[inner sep=0pt] at (\xsep, 0*\ysep) {\includegraphics[trim=0 1.4cm 0 1.4cm,clip,width=\rsize\textwidth]{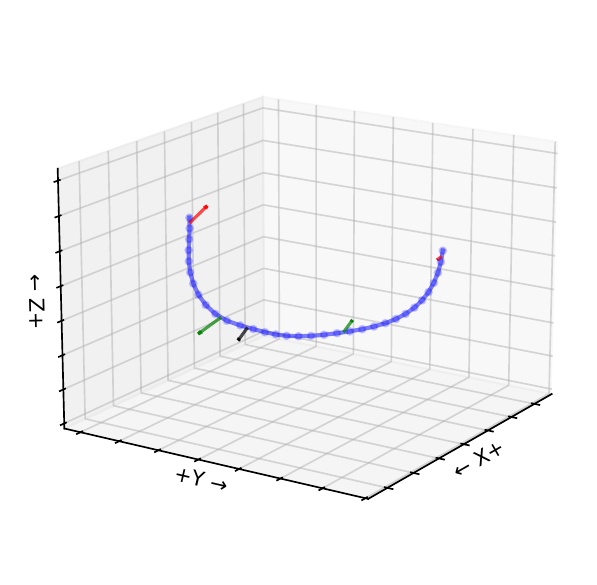}};
		
		\node[inner sep=0pt] at (-\xsep, -1*\ysep) {\fbox{\includegraphics[width=\lsize\textwidth]{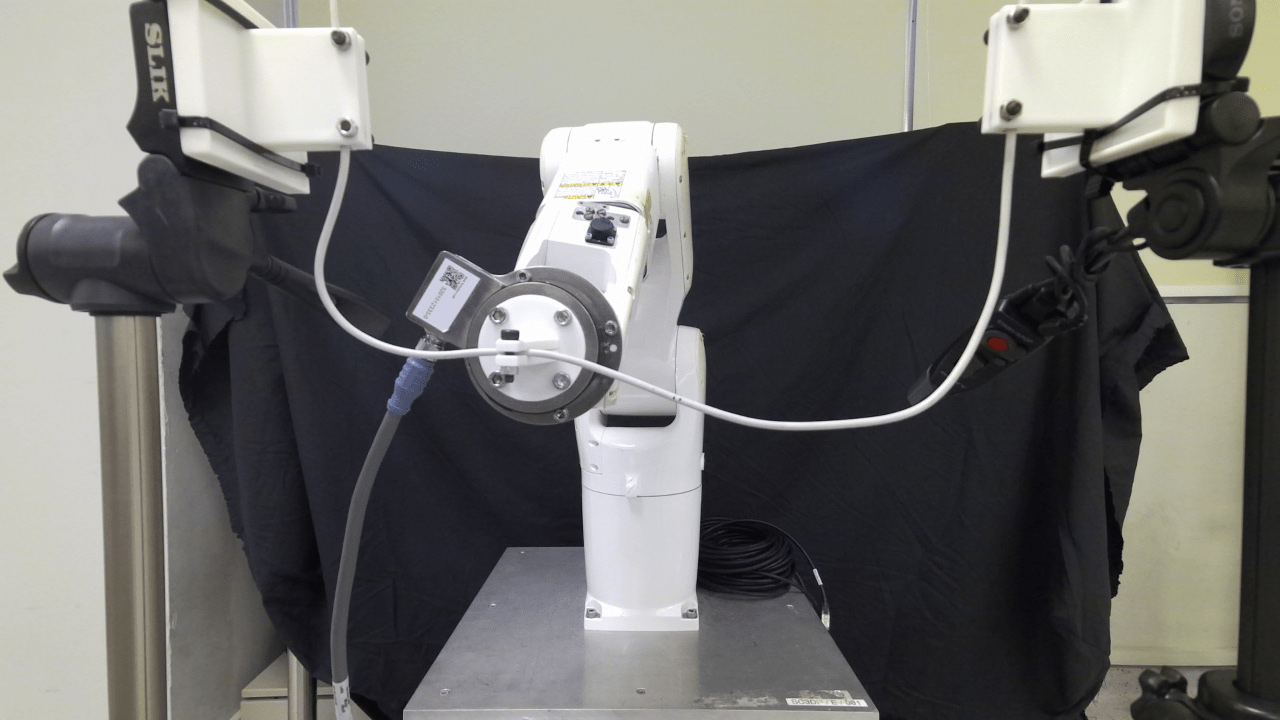}}};
		\node[inner sep=0pt] at (\xsep, -1*\ysep) {\includegraphics[trim=0 1.4cm 0 1.4cm,clip,width=\rsize\textwidth]{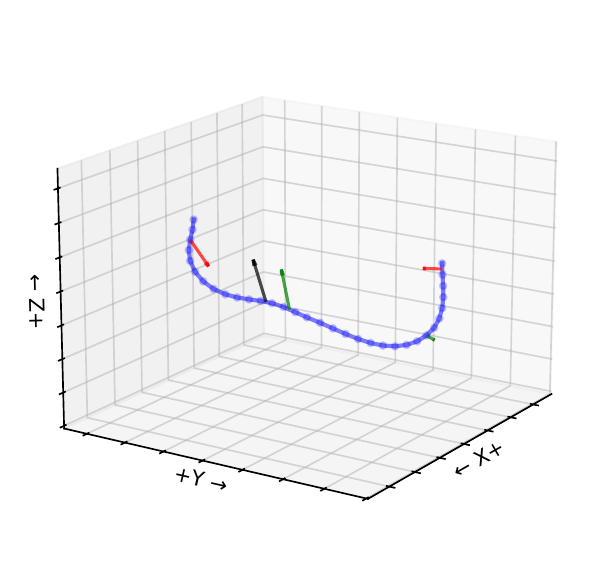}};
		
		\node[inner sep=0pt] at (-\xsep, -2*\ysep) {\fbox{\includegraphics[width=\lsize\textwidth]{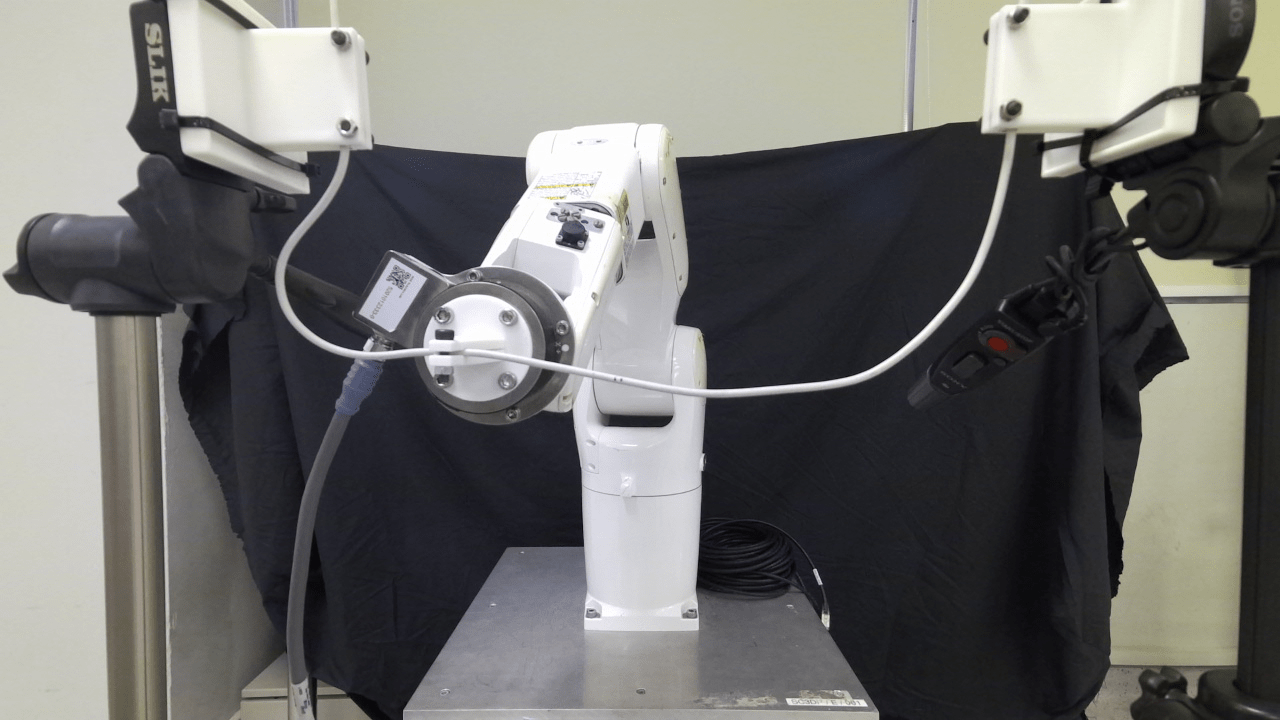}}};
		\node[inner sep=0pt] at (\xsep, -2*\ysep) {\includegraphics[trim=0 1.4cm 0 1.4cm,clip,width=\rsize\textwidth]{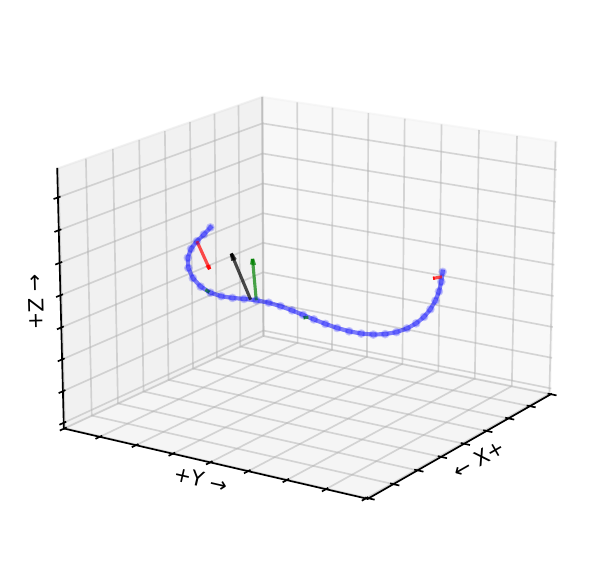}};
		
		\node[inner sep=0pt] at (-\xsep, -3*\ysep) {\fbox{\includegraphics[width=\lsize\textwidth]{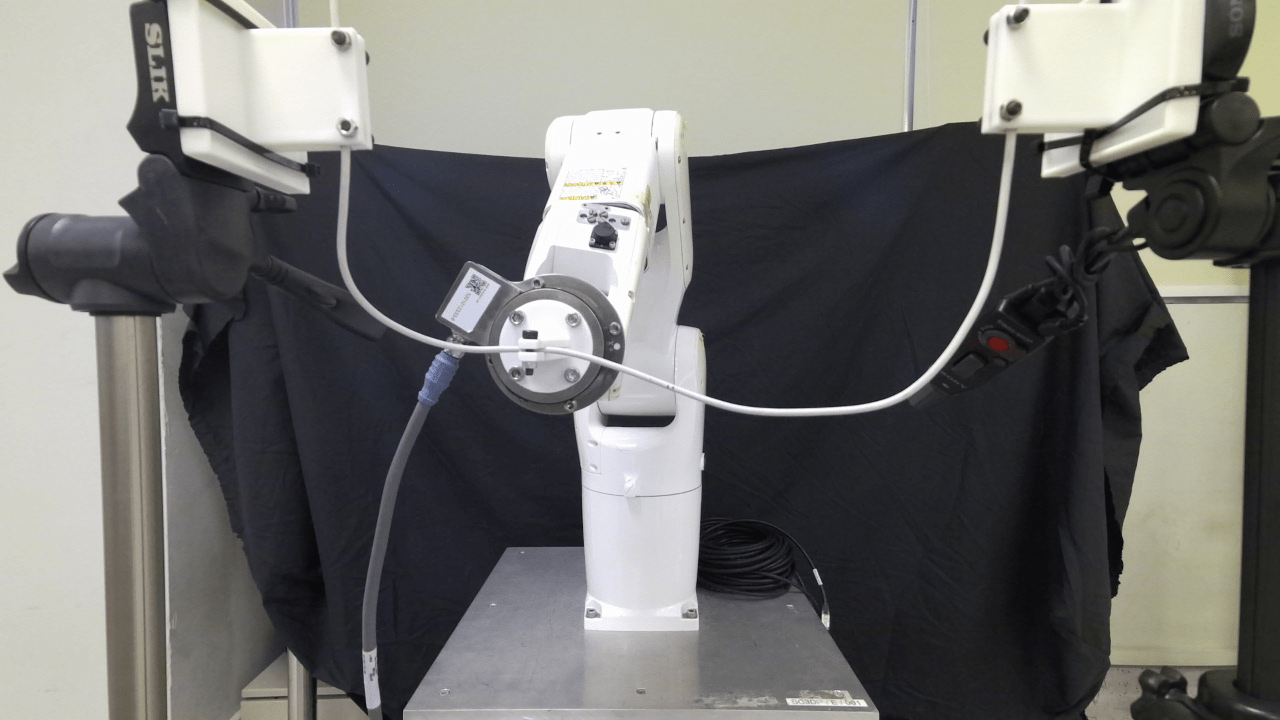}}};
		\node[inner sep=0pt] at (\xsep, -3*\ysep) {\includegraphics[trim=0 1.4cm 0 1.4cm,clip,width=\rsize\textwidth]{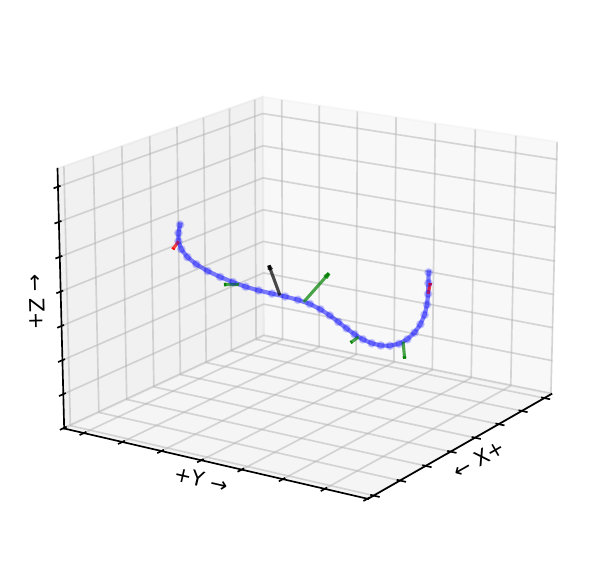}};
	\end{tikzpicture}
	\caption[Visual Results of Force Estimation for Off-centered Displacement]{Visual results of force estimation for experiment \textbf{B} where the wire was clamped at both ends and attached to the robot end-effector at a $\SI{7}{cm}$ offset along the wire length from its center.}
	\label{Fig:s2fexpB_vis}
\end{figure}

For the second part (\textbf{B}), the wire was clamped to the robot end-effector at a $\SI{7}{cm}$ offset along the wire length from its center ($\SI{18}{cm}$ along its length from the left end) in its equilibrium position and moved into 4 different positions with pure translational motion.
The actual and estimated force vectors for experiment \textbf{B} are shown in \Cref{Fig:s2fexpB_vis}. Overall quantitative results for both experiments (\textbf{A} and \textbf{B}) comparing the actual ($\mathbf{F}_{\text{act}}$) and estimated ($\mathbf{F}_{\text{est}}$) forces are shown in \Cref{Table:s2f_force_data}. To evaluate the accuracy of the estimated force \( \mathbf{F}_{\text{est}} \) compared to the actual force \( \mathbf{F}_{\text{act}} \), we computed several error metrics. To assess directional accuracy, we computed the \emph{angle difference} between the vectors. The \emph{relative L2 error} normalizes the L2 error by the magnitude of the actual force, yielding $\frac{ \left\| \mathbf{F}_{\text{est}} - \mathbf{F}_{\text{act}} \right\|_2 }{ \left\| \mathbf{F}_{\text{act}} \right\|_2 + \varepsilon },
$ where $\varepsilon$ is a small constant added to avoid division by zero. Lastly, to evaluate the spatial accuracy of the estimated point of application, we computed the \emph{position difference} between the estimated and actual contact points as $\left\| \mathbf{p}_{\text{est}} - \mathbf{p}_{\text{act}} \right\|_2,$ where $\mathbf{p}$ is the position at which the force acts.

\begin{table}[!htbp]
	\vspace{0.5cm}
	\caption{Quantitative Results Comparison for Force Estimation}
	\vspace{-0.5cm}
	\label{Table:s2f_force_data}
	\begin{center}
		\resizebox{\columnwidth}{!}{%
			\begin{tabular}{|>{\centering}m{15mm}|p{23mm}|p{23mm}|p{23mm}|}
				\hline
				\rule{0pt}{3ex} Experiment & Relative L2 Err. (N) & Angle Diff. ($^\circ$) & Position Diff. (mm)\\
				\hline
				\textbf{A1} & 0.2521 & 0.2225 & 8.242\\
				\hline
				\textbf{A2} & 0.1364 & 0.1176 & 0.0006\\
				\hline
				\textbf{A3} & 0.1522 & 0.1452 & 24.93\\
				\hline
				\textbf{A4} & 0.8009 & 0.8887 & 8.278\\
				\hline
				\textbf{A5} & 0.6218 & 0.2091 & 8.422\\
				\hline
				\textbf{A6} & 0.4223 & 0.0367 & 32.31\\
				\hline
				\textbf{B1} & 1.0491 & 1.1323 & 33.45\\
				\hline
				\textbf{B2} & 0.1281 & 0.1103 & 33.32\\
				\hline
				\textbf{B3} & 0.4021 & 0.4133 & 7.494\\
				\hline
				\textbf{B4} & 0.9419 & 0.9235 & 30.67\\
				\hline
			\end{tabular}
		}
	\end{center}
\end{table}

\subsection{Discussion}
In scenarios \textbf{A4}, \textbf{A5}, \textbf{B1}, and \textbf{B4}, additional erroneous forces were detected. For our analysis, we only compared the largest magnitude force with the actual force sensed. The \emph{relative L2 error} and \emph{angle difference} were generally lower for single direction wire displacements, with the exception of \textbf{B1}. Along with \textbf{A4}, \textbf{A6}, and \textbf{B4}, we found that displacement and forces in the $x$-direction were not easily detected with our method and caused relatively large relative L2 errors and angle differences. For the cases of \textbf{A4} and \textbf{B4}, the algorithm seemed to dissect the forces and identify additional force disturbances (other green arrows) which have an $x$-direction contribution, but these were not considered in our analysis.

Interestingly, these forces which were difficult to detect were those that act nearly tangentially to the wire. That could be a hint that the algorithm in its current state is not suited to accurately determine tangential forces due to the sensitivity of calculated tangential forces to the error in positions and orientations of the discretized wire pieces, and  wire physical properties. We speculate that these errors likely result from two main causes: the wire not having purely elastic deformation properties and imperfect determination of wire position using the depth camera. The latter of which can be solved with improved detection capabilities. The former proves to be a more complex issue and could directly affect our results as the algorithm might recognize internal torque in the wire where it is not actually present, requiring further intense investigations into modeling of plastic deformations in DLOs.

Estimations on the position of force application were all below $\SI{40}{mm}$ and experienced better results for experiment \textbf{A} than \textbf{B}. This could be due to a displacement at the midpoint of the wire \textbf{A} producing more deformation in the wire which the algorithm is sensitive to.

\section{Conclusion}
In this work, we introduced an algorithm to estimate external forces acting on an elastic rod based on its observed shape. By deriving consistency conditions from static force-torque balance equations, we identified undisturbed sections of the rod where no external interactions occur. Leveraging this classification and internal stiffness torque models from the DER formulation, we solved for the direction and magnitude of external forces. Experimental results validated the method’s effectiveness by comparing the estimated forces against ground truth readings from physical sensors during robot-wire interactions.

\subsection{Limitations and Future Work}
Through our experiments, we assume the wire behaves in a purely elastic way which is not an accurate model for its true behavior. The real wire experiences plastic deformation as can be seen from the undisturbed kinks sometimes found in the wire when recovering from large deformation. Although this property is partially hidden when smoothing the visually detected DLO, it definitely affects the accuracy of our results. Past works aimed at manipulating wires with such properties use learning techniques~\cite{laezza2021learning,matl2021deformable} or model the plasticity directly with known physical properties~\cite{terzopoulos1988deformable}.

Wire detection using the Azure Kinect’s time-of-flight infrared depth sensor is prone to significant noise, particularly in regions where the wire is in contact with or occluded by other objects. This noise adversely affects the accuracy of shape and force estimation. The proposed method would benefit significantly from a higher-resolution, lower-noise wire detection approach. This remains an active area of research within the field of deformable linear object perception~\cite{caporali2023rt,xiang2023trackdlo,lv2022learning,zhaole2023robust}.


%

\addtolength{\textheight}{-0cm}   

\bibliographystyle{ieeetran}
\bibliography{ref.bib}

\end{document}